\documentclass[conference]{IEEEtran}
\IEEEoverridecommandlockouts
\usepackage{cite}
\usepackage{amsmath,amssymb,amsfonts}
\usepackage{algorithmic}
\usepackage{graphicx}
\usepackage{textcomp}
\usepackage{xcolor}

\usepackage{wrapfig}
\usepackage{graphics} 
\usepackage{tabularx}
\usepackage{epsfig} 
\usepackage{multirow}
\usepackage{verbatim} 
\usepackage{subcaption}
\usepackage{hyperref}
\usepackage{url}

\def\BibTeX{{\rm B\kern-.05em{\sc i\kern-.025em b}\kern-.08em
    T\kern-.1667em\lower.7ex\hbox{E}\kern-.125emX}}
\begin{document}

\title{Neural Task Success Classifiers for Robotic Manipulation from Few Real Demonstrations
}

\author{\IEEEauthorblockN{Abdalkarim Mohtasib}
\IEEEauthorblockA{\textit{School of Computer Science} \\ \\
\textit{University of Lincoln}\\
Lincoln, UK \\
amohtasib@lincoln.ac.uk}
\and
\IEEEauthorblockN{Amir Ghalamzan E.}
\IEEEauthorblockA{\textit{Lincoln Institute for Agri-Food} \\ Technology \\
\textit{University of Lincoln}\\
Lincoln, UK}
\and
\IEEEauthorblockN{Nicola Bellotto}
\IEEEauthorblockA{\textit{School of Computer Science} \\ \\
\textit{University of Lincoln}\\
Lincoln, UK}
\and
\IEEEauthorblockN{Heriberto Cuay\'ahuitl}
\IEEEauthorblockA{\textit{School of Computer Science} \\ \\
\textit{University of Lincoln}\\
Lincoln, UK}
}
\maketitle

\begin{abstract}
Robots learning a new manipulation task from a small amount of demonstrations are increasingly demanded in different workspaces. A classifier model assessing the quality of actions can predict the successful completion of a task, which can be used by intelligent agents for action-selection. 
This paper presents a novel classifier that learns to classify task completion only from a few demonstrations. We carry out a comprehensive comparison of different neural classifiers, e.g. fully connected-based, fully convolutional-based, sequence2sequence-based, and domain adaptation-based classification. We also present a new dataset including five robot manipulation tasks, which is publicly available. We compared the performances of our novel classifier and the existing models using our dataset and the MIME dataset. The results suggest domain adaptation and timing-based features improve success prediction. Our novel model, i.e. fully convolutional neural network with domain adaptation and timing features, achieves an average classification accuracy of 97.3\% and 95.5\% across tasks in both datasets whereas state-of-the-art classifiers without domain adaptation and timing-features only achieve 82.4\% and 90.3\%, respectively.
\end{abstract}


\section{INTRODUCTION}
Large amounts of tasks are carried out in our daily lives, in large variation either due to the way they are done or to the environments where they are executed, and robots are expected to learn some of these tasks to be able to assist humans. In order for this to happen, robots should be able to learn tasks in an autonomous fashion as opposed to hard-coding all robot skills. Humans are able to learn new skills rather rapidly and, in many cases (not always because some tasks are difficult to master), using only a few examples. Arguably, robots should be endowed with similar or even better learning abilities to be able to acquire new tasks quickly and efficiently. Being able to identify the task goal and to measure task success is the first key aspect for robots to acquire new tasks autonomously. 

\begin{figure*}[t]
\centerline{\includegraphics[scale=0.55]{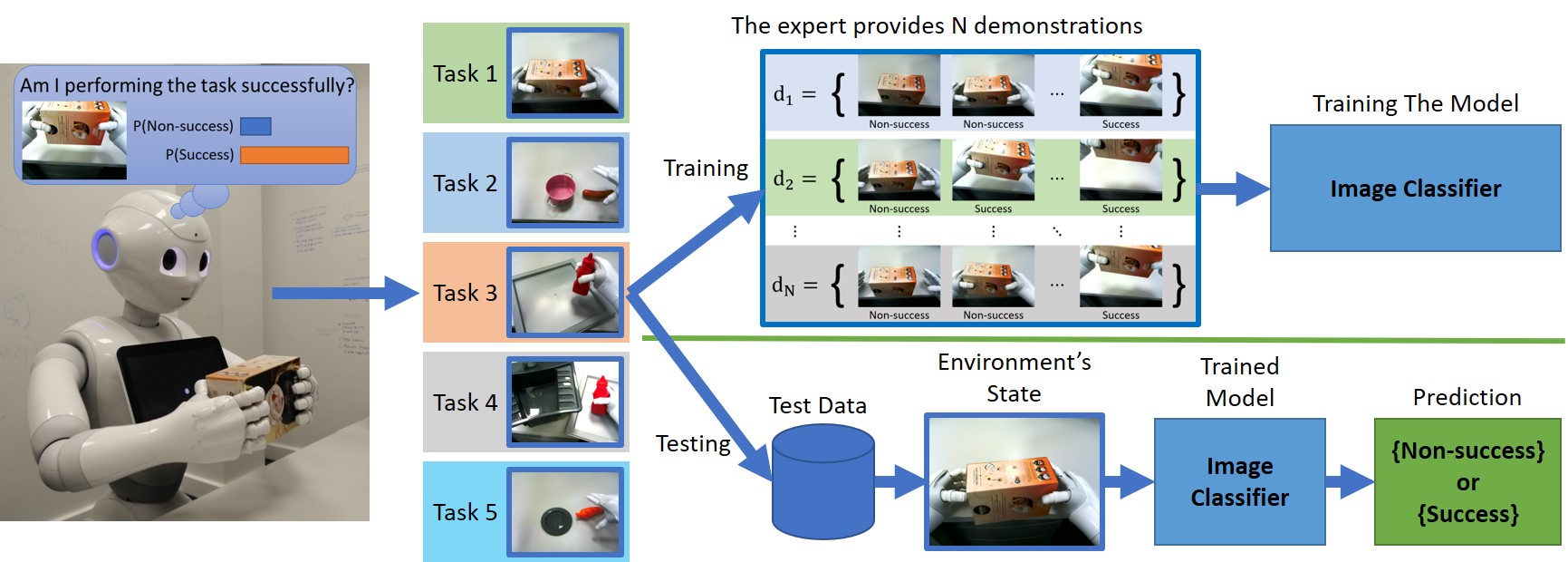}}
\caption{Illustration of the targeted robot learning scenario}
\label{fig2}
\end{figure*}

Reward functions in robot learning play a major role in measuring task success in order to numerically reward the behaviour of robots \cite{KoberBP13}. The use of onboard robot sensors only---without relying on any other external sensors---makes the problem of measuring the success of tasks even harder. In this context, this paper focuses on training success classifiers (also referred to as 'goal classifiers') for measuring the levels of success in robotic manipulation tasks from only a few (as opposed to many) human demonstrations. 
The idea of training a success classifier in a new task using only a few demonstrations with high accuracy is still a challenging research problem. In this work, we study such a problem via the scenario illustrated in Fig.~\ref{fig2}. 

Our main contribution in this paper is a comprehensive comparison of  different neural architectures for task success classification, based on feedforward neural networks, fully convolutional neural networks, sequence2sequence classifiers, domain adaptation, and a novel combination of them. This study was carried out using a newly proposed dataset of human-robot demonstrations in the Kitchen domain as well as an existing dataset of demonstrations \cite{SharmaMPG18} using a variety of manipulation tasks including stacking, placing, opening, closing, rolling, pushing, pulling, and rotating objects. 
\section{Related Work}
The topic of reward learning for trainable robots has been studied in several ways in the academic literature. \cite{lopes2009active, cui2018active, daniel2014active, sadigh2017active} have studied how to solve this problem using Active Learning, which involves querying the expert user for labelling some trajectories or environment states. Similar approaches have combined the use of active learning together with Inverse Reinforcement Learning (IRL) \cite{lopes2009active, cui2018active}, where IRL is used to estimate and optimise a reward function by learning from demonstration in Markov Decision Processes (MDPs) \cite{ng2000algorithms}. Meta-Learning has also been used by robots that learn to acquire new tasks via knowledge transfer from a large set of pre-learned tasks \cite{xie2018few, james2018task} to a new task. While meta-learning approaches have achieved promising results in learning the goals of new tasks from few demonstrations, they unfortunately require a very large number of training examples to train the meta-learner. Furthermore, the `goal classifier' approach has been adopted by different researchers \cite{singh2019end,vecerik2019practical,pinto2016supersizing}. It uses an image-based classifier to predict whether an environment state (an image of the task at time $t$) represents a success or non-success of the executed task. Those goal classifiers usually require a substantial number of training examples, as well as an extensive effort from human demonstrators to be able to successfully train task success predictors---this means that their effective and efficient training must be studied further.

    Previous works are limited by relying on engineered reward functions \cite{vevcerik2017leveraging,james20163d} and by using additional sensors (i.e. in addition to the onboard sensors) to estimate task success \cite{yahya2017collective,yamaguchi2015pouring,schenck2017visual}. Other previous works have investigated the idea of learning the reward function from demonstration data \cite{abbeel2004apprenticeship,finn2016guided,ziebart2008maximum,ratliff2006maximum,edwards2017perceptual,lopes2009active,syed2008apprenticeship,brown2019risk,cohn2011comparing} mainly by using Inverse Reinforcement Learning---but they are unfortunately very data intensive. Other research projects have used active learning to query the expert for labelling uncertain states~\cite{singh2019end} or execution trajectories  \cite{daniel2014active} to learn the task rewards---but their practical deployment to end users is unclear. Our work differs from previous ones in automatically inducing reward functions from raw data, in studying data-efficient methods for their practical application, and in the use of onboard sensors only (a 2D camera in our case) without any external sensors in the environment.
	
	The main related works to ours that have used task success classifiers are mostly based on Convolutional Neural Networks (CNNs) \cite{singh2019end,vecerik2019practical,pinto2016supersizing,tung2018reward,levine2018learning,fu2018variational,xie2018few,edwards2017cross,sermanet2016unsupervised}, 
	but they have not exploited sequential and timing aspects of robotic manipulation tasks for inducing their decisions. This paper investigates the effects of using timing information in manipulation tasks and sequential behaviour in an attempt to improve the performance of task success classifiers.
	In addition, domain adaptation techniques \cite{singh2019end,vecerik2019practical,pinto2016supersizing,tung2018reward,levine2018learning,fu2018variational,xie2018few,edwards2017cross,sermanet2016unsupervised} have not been applied to reward learning. Domain adaptation refers to the case where what has been learnt in one domain (demonstrations in our case) is exploited to improve generalisation in another domain (unseen conditions in our case) \cite{goodfellow2016deep}. This paper investigates the effects of using domain adaptation techniques to improve the performance of reward predictors casted as success classifiers. 
	Thus, the context of this paper is to develop trainable reward models (as opposed to hard-coded) that can be used to accurately measure task success in robotic manipulation tasks. Future works can use such models as a part of numerical rewards required for robot learning systems. The code, models, and data produced as part of this paper are publicly available on GitHub\footnote{\url{https://Mohtasib.github.io/RewardLearning/}}.
	
    

\section{Research Methods}

\subsection{Problem Definition}

We consider a success classifier $g=f(s)$, where $s$ is the environment state (an image or sequence of images from the robot's onboard 2D camera), and \(g\in\left[0,1\right]\) is the probability of having achieved the task in state $s$. This classifier can be used as a reward function for robot learning systems. The aim is to train $f(s)$ for a new manipulation task \(M_n\) from $N$ demonstrations by updating the parameters of $g$ to minimize \(\sum{\mathcal{L}(f(s_i),y_i)}\), where \(\mathcal{L}\) is the classification loss (cross entropy loss and mean square error in our case). We define \(D_n=\left\{d_1,d_2,\cdots,d_N\right\}\) as the demonstrations dataset for the new task \(M_n\), and each demonstration is defined as a set of states $s$ and their labels $y$ as follows: \(d_i=\left\{\left(s_1,y_1\right),\left(s_2,y_2\right),\cdots,\left(s_j,y_j\right)\right\}_i\). The label $y$ is $0$ if the state $s$ represents a Non-success in the task being executed, and $1$ if the state $s$ represents a Success in the task being executed. The research question that our study aims to answer is: {\it Can a task success classifier be trained effectively from a very small number of demonstrations (e.g. five)?}


    \begin{table*}[!ht]
    \caption{Example training and test images in the Kitchen and MIME datasets, K$_i$ and M$_j$, respectively} 
    \begin{center}
    \begin{tabular}{c|c|c|c}

    \textbf{Dataset} & \textbf{Task} & \textbf{Training Examples} & \textbf{Test Examples} \\ \hline 
    \multirow{17}{*}{Kitchen} & 
    \multirow{3}{*}{K$_1$}        &      
    \multirow{17}{*}{\includegraphics[scale=0.430]{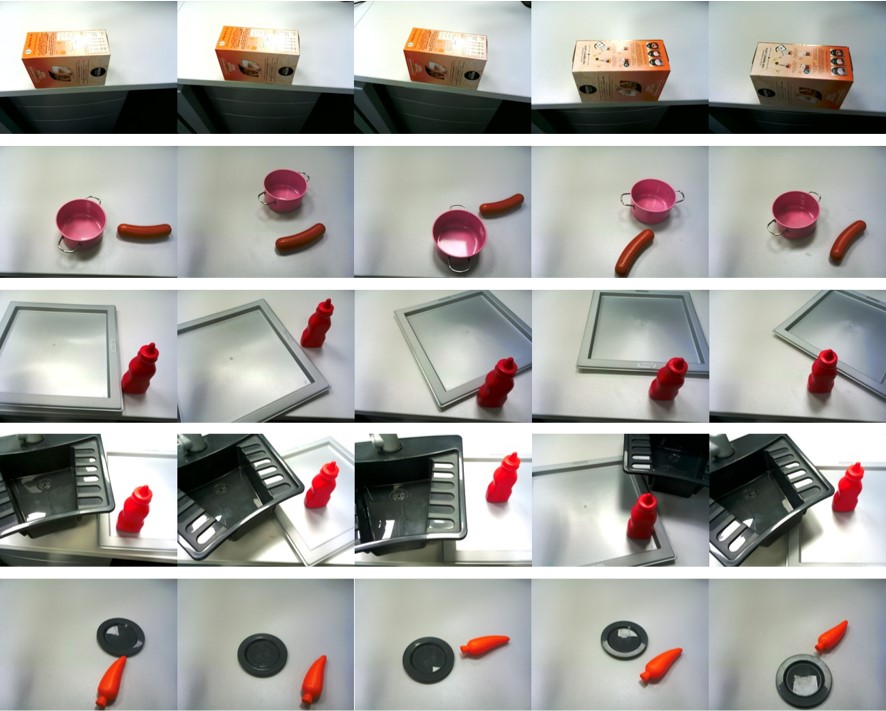}}   &        \multirow{17}{*}{\includegraphics[scale=0.430]{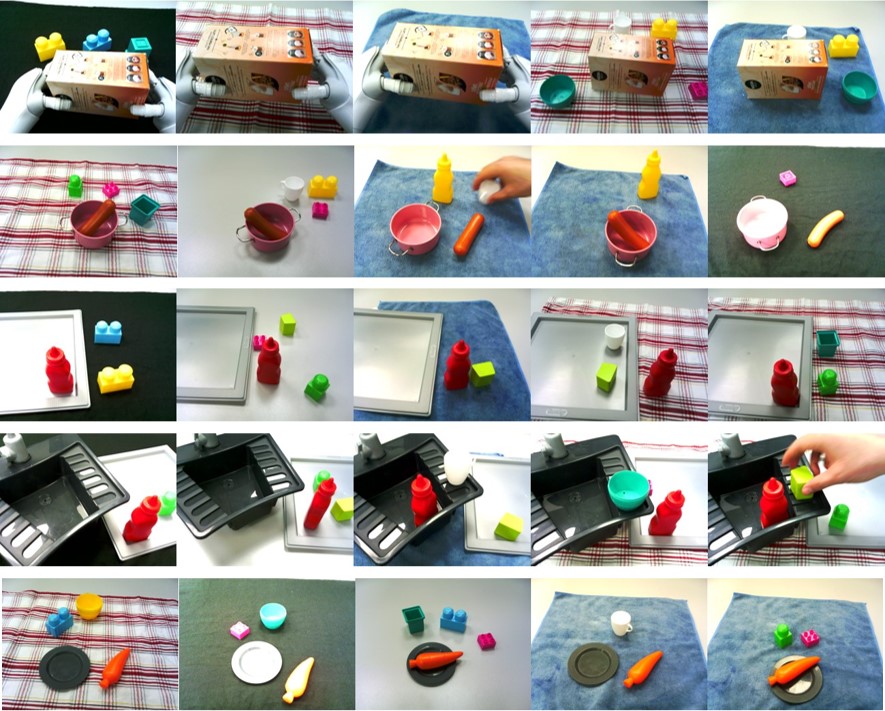}} \\  &   &  \\  &   &  \\
    & \multirow{4}{*}{K$_2$} &    &  \\  &    &  \\  &    &  \\&    &  \\
    & \multirow{3}{*}{K$_3$} &    &  \\  &    &  \\  &    &  \\
    & \multirow{4}{*}{K$_4$} &    &  \\  &    &  \\  &    &  \\&    &  \\
    & \multirow{3}{*}{K$_5$} &    &  \\  &    &  \\  &    &  \\
    \hline \hline
    \multirow{30}{*}{MIME} & 
    \multirow{3}{*}{M$_1$}        &      
    \multirow{30}{*}{\includegraphics[scale=0.430]{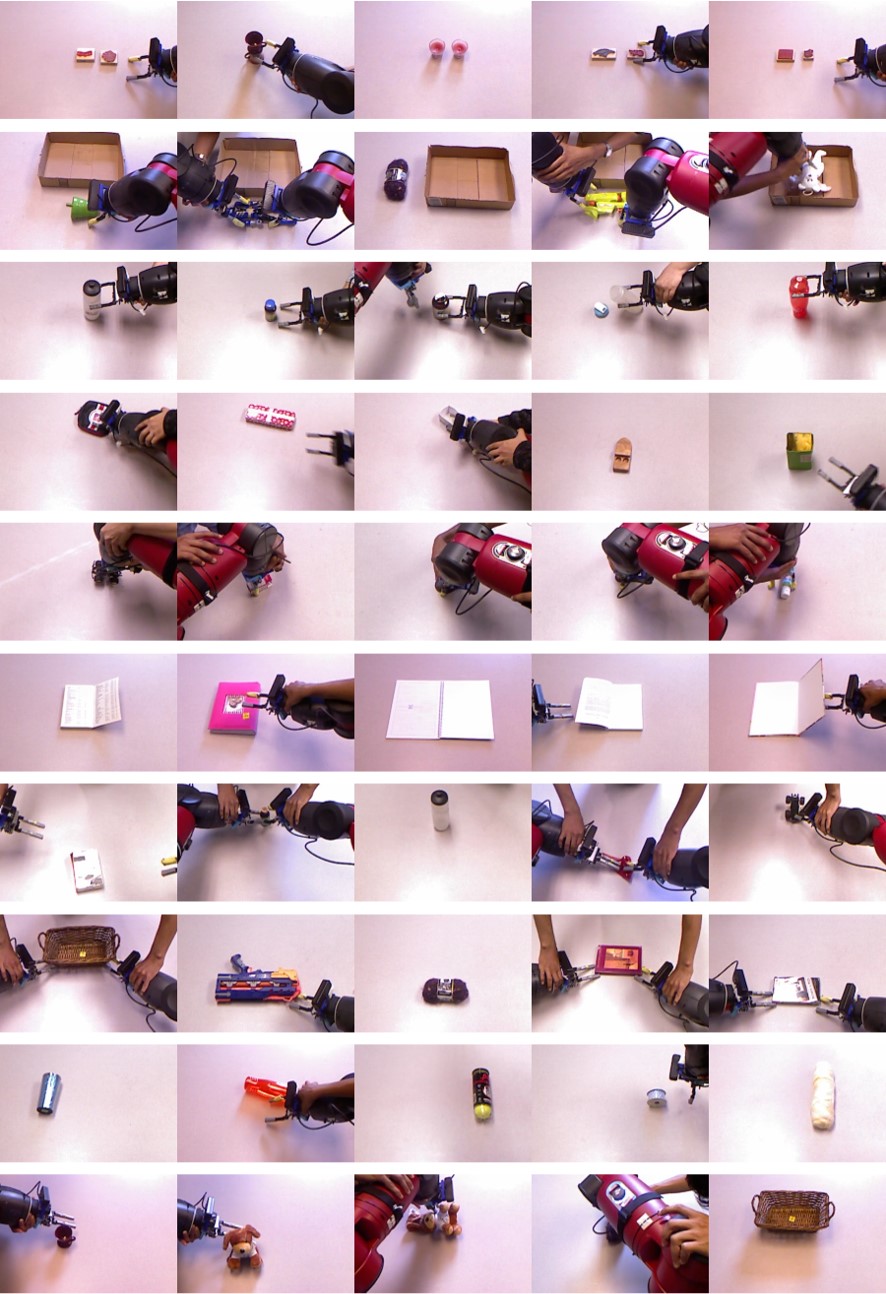}}   &        \multirow{30}{*}{\includegraphics[scale=0.430]{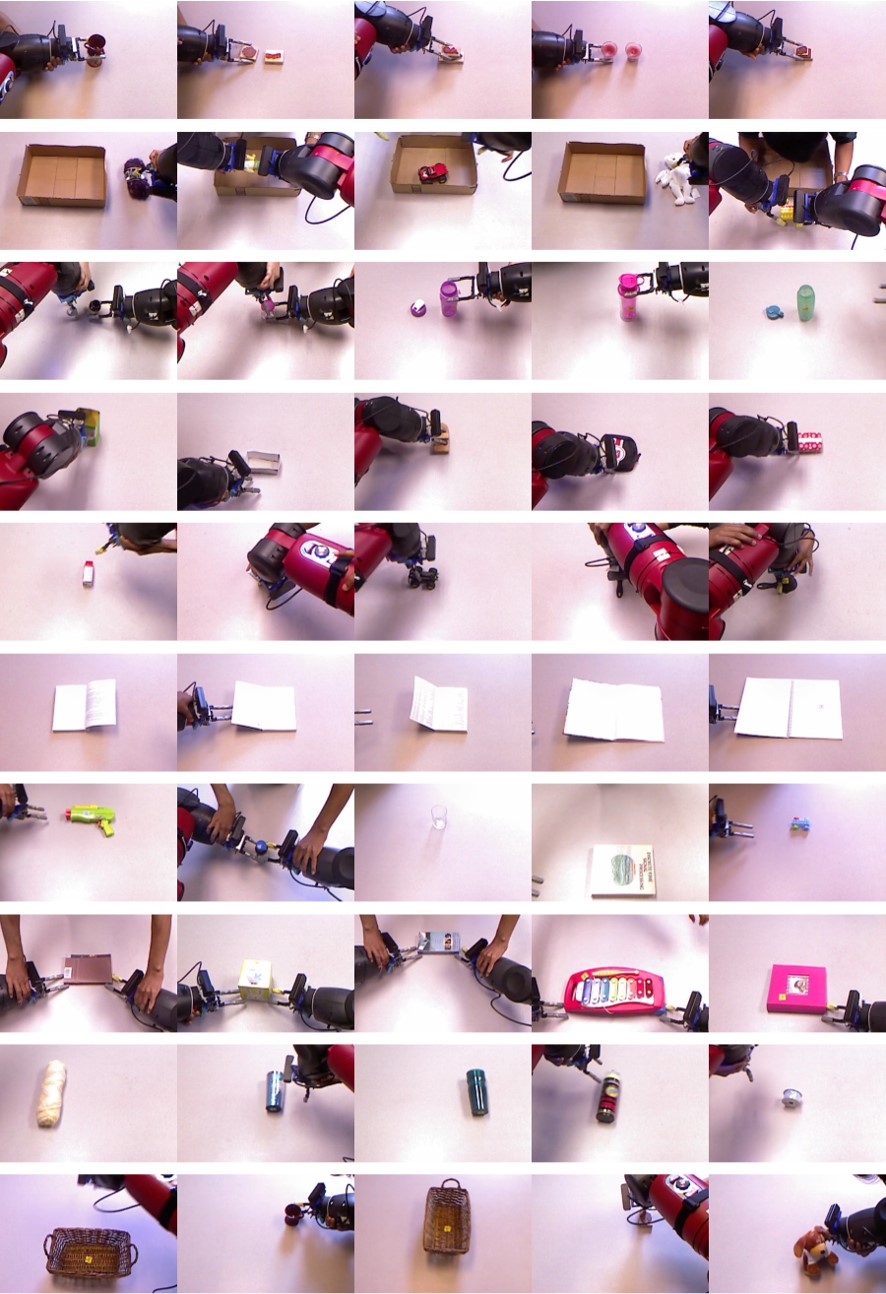}} \\  &   &  \\  &   &  \\
    & \multirow{3}{*}{M$_2$} &    &  \\  &    &  \\  &    &  \\
    & \multirow{3}{*}{M$_3$} &    &  \\  &    &  \\  &    &  \\
    & \multirow{3}{*}{M$_4$} &    &  \\  &    &  \\  &    &  \\
    & \multirow{3}{*}{M$_5$} &    &  \\  &    &  \\  &    &  \\
    & \multirow{3}{*}{M$_6$} &    &  \\  &    &  \\  &    &  \\
    & \multirow{3}{*}{M$_7$} &    &  \\  &    &  \\  &    &  \\
    & \multirow{3}{*}{M$_8$} &    &  \\  &    &  \\  &    &  \\
    & \multirow{3}{*}{M$_9$} &    &  \\  &    &  \\  &    &  \\
    & \multirow{3}{*}{M$_{10}$} &    &  \\  &    &  \\  &    &  \\

    \end{tabular}
    \label{tasks_datasets}
    \end{center}
    \end{table*}

\subsection{Datasets}
\label{dataset}
Our proposed dataset of human-robot demonstrations was collected using the Pepper Robot\footnote{\url{https://www.softbankrobotics.com/emea/en/pepper}}. In this dataset, a human demonstrator performed kinesthetically the manipulation tasks for the robot by grabbing the robot hands and performing tasks in the Kitchen domain. During the task demonstrations, color images from the robot's 2D camera (size $320 \times 240 \times 3$) were recorded. When a demonstration task is completed, the demonstrator touched a tactile sensor on the robot head to signal task success. In this way and for practical purposes, any data collected before the tactile sensor was touched is labelled as {\it Non-success}, and any data collected after that is labelled as {\it Success}. The robot collects data at a sampling rate of 10 samples/sec. Example images captured using the Pepper robot's 2D camera for grasping, lifting, picking up and placing kitchen objects are illustrated in Fig.~\ref{fig2}. 

    \begin{table}
    \caption{Training and test examples in the Kitchen dataset, per task, containing Non-Success (NS) and Success (S) images}
    \begin{center}
    \begin{tabular}{|c|l|c|c|c|}
    \hline
  \multirow{2}{*}{\textbf{\#}} &  \multirow{2}{*}{\textbf{Description}}      & {\textbf{Training}} & {\textbf{Test}} \\ 
                                        &     & NS / S                   & NS / S                 \\ \hline
K$_1$ & Grasp \& lift a box                       & 1281 / 194                 & 74 / 85                \\ \hline
K$_2$ & Pick up a sausage \& place it in a cooker & 946 / 148                 & 138 / 86                \\ \hline
K$_3$ & Pick up a ketchup \& place it on a tray   & 1533 / 199                 & 232 / 67                \\ \hline
K$_4$ & Pick up a ketchup \& place it in a sink   & 1093 / 264                 & 162 / 123               \\ \hline
K$_5$ & Pick up a carrot \& place it on a plate   & 1338 / 257                 & 311 / 69                \\ \hline
    \end{tabular}
    \label{tasks_tab1}
    \end{center}
    \end{table}

    \begin{table}
    \caption{Summary of tasks in the MIME dataset}
    \begin{center}
    \begin{tabular}{|c|l|c|c|c|}
    \hline
  \textbf{\#} &  \textbf{Description}      & {\textbf{Training}} & {\textbf{Test}} \\ \hline
M$_1$ & Stack                                   & 888 / 103                 & 828 / 96                  \\ \hline
M$_2$ & Place objects in box                    & 1718 / 360                & 1576 / 283                \\ \hline
M$_3$ & Open bottles                            & 1755 / 133                & 1594 / 135                \\ \hline
M$_4$ & Push (Single hand)                      & 323 / 60                  & 301 / 80                  \\ \hline
M$_5$ & Rotate                                  & 808 / 135                 & 778 / 130                 \\ \hline
M$_6$ & Close Book                             & 604 / 177                 & 555 / 168                 \\ \hline
M$_7$ & Pull (Two hands)                       & 1538 / 145                & 1515 / 151                \\ \hline
M$_8$ & Push (Two hands)                       & 1816 / 384                & 1879 / 333                \\ \hline
M$_9$ & Roll                                   & 489 / 100                 & 377 / 130                 \\ \hline
M$_{10}$ & Pull (Single hand)                     & 544 / 94                  & 466 / 83                  \\ \hline
    \end{tabular}
    \label{tasks_tab2}
    \end{center}
    \end{table}    
    
    \begin{figure*}
    \centerline{\includegraphics[scale=0.58]{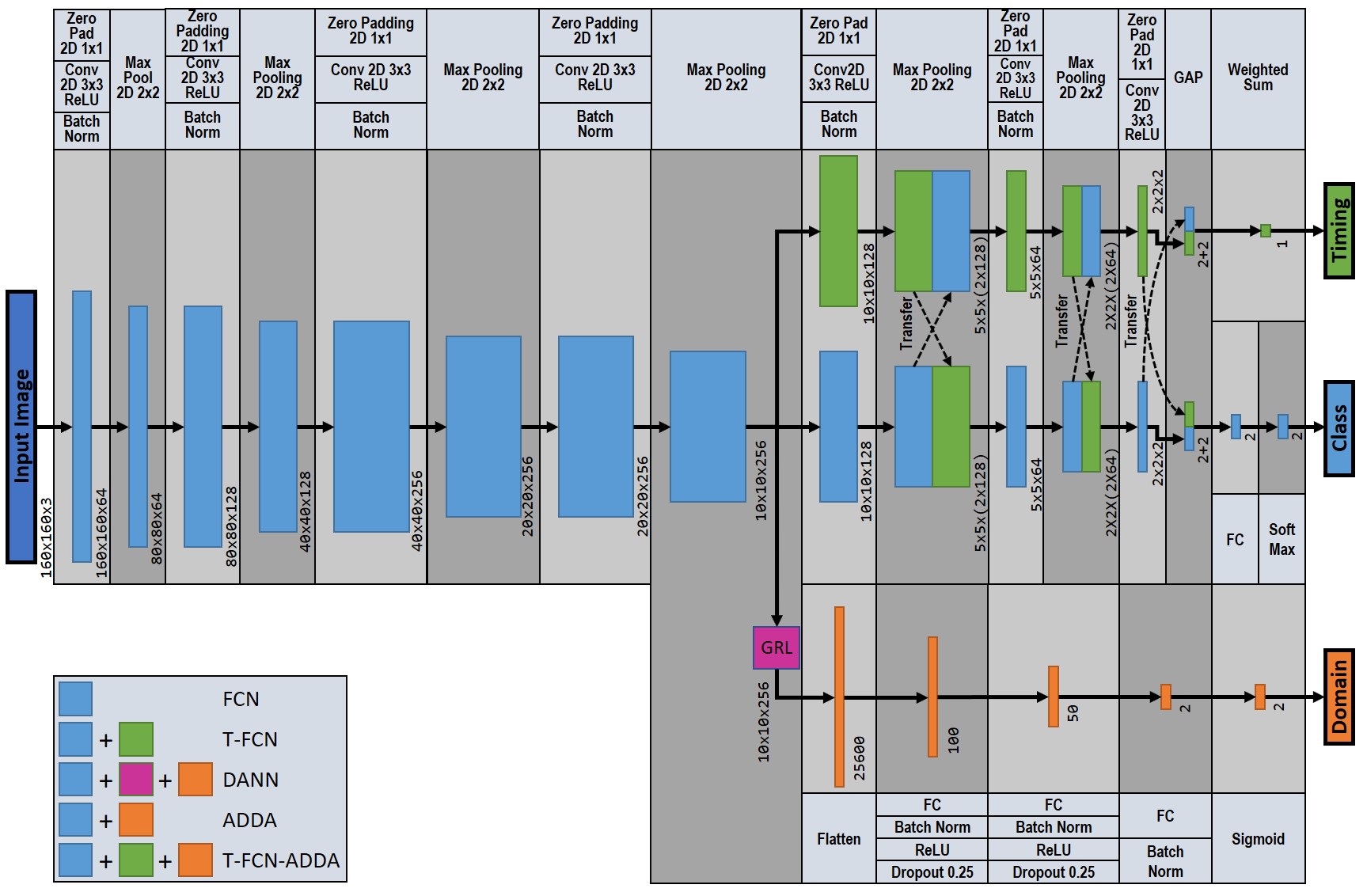}}
    \caption{CNN-based neural architectures for task success classification}
    \label{models_fig2}
\end{figure*}

While demonstration data of five different robotic manipulation tasks were collected (see list of tasks in Table~\ref{tasks_tab1}), each of these demonstrations was carried out at a different speed and with objects positions initialised randomly. The randomisation was limited to the workspace of the robot's arms and to the view of the robot's camera (see K$_i$ examples in Table~\ref{tasks_datasets}). 
    
The MIME (Multiple Interactions Made Easy) dataset has been used for learning from demonstrations in multiple tasks \cite{SharmaMPG18}. It consists of kinesthetic trajectories and videos of human demonstrations collected using the Baxter robot ({\url{https://en.wikipedia.org/wiki/Baxter_(robot)}}), i.e. human demonstrations and their corresponding robot demonstrations. To collect this dataset, a set of human demonstrators were trained to handle the robot before performing the demonstration task. Every human demonstrator provided multiple demonstrations for each task using different objects --filtering out incorrect cases. The dataset contains in total 8260 demonstrations for 20 tasks, from which we extracted 10 demonstrations (5 for training and 5 for testing) for the 10 M$_j$ tasks shown in Table~\ref{tasks_datasets}. These 10 tasks and their demonstrations have been selected randomly.


\subsection{Model Architectures}
\label{models}
The following neural architectures are studied in this paper.
\vskip-0pt
\begin{figure}[!ht]
    \centerline{\includegraphics[scale=0.55]{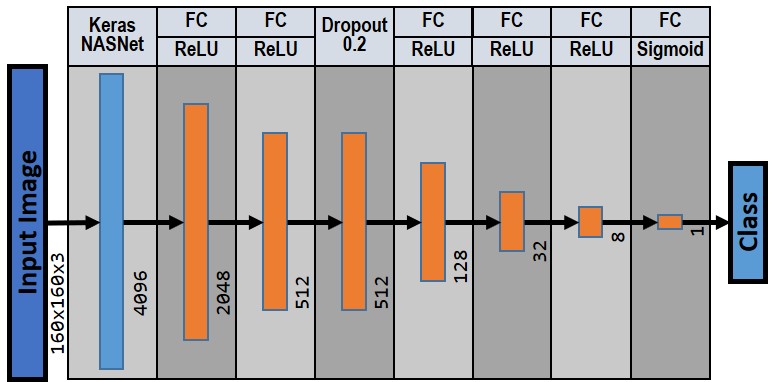}}
    \caption{NASNet-based neural architecture}
    \label{models_fig1}
\end{figure}

\begin{itemize}
\item {\it Fully Connected Neural Net} ({\bf NASNet}): This baseline is NASNet-based \cite{zoph2018learning} with pre-trained weights on ImageNet \cite{imagenet_cvpr09}. It has the best classification performance compared to the other models implemented in the Keras API~\cite{chollet2015keras}. In our model, features extracted from NASNet are passed to a Feed-Forward Network consisting of six fully-connected layers as illustrated in Fig.~\ref{models_fig1}.
\vskip5pt
\item {\it Fully Convolutional Neural Net} ({\bf FCN}): This second baseline is very similar to the CNN-based models that have been used in literature as  success classifiers  \cite{singh2019end,vecerik2019practical,tung2018reward,levine2018learning,fu2018variational,xie2018few,edwards2017cross,sermanet2016unsupervised}. This network is the core of other architectures implemented in this paper. The inputs to FCN are (160 × 160 × 3) resized images from the robot’s 2D camera, followed by six main convolutional blocks and one convolutional layer, see Fig.~\ref{models_fig2}.
\vskip5pt
\item {\it Time-Based Fully Convolutional Neural Net} ({\bf T-FCN}): This architecture extends FCN with two paths and features (shared in between): one is the classification path, the other is a timing path that predicts the proportion of task completion (a regressor), see Fig.~\ref{models_fig2}. The task completion proportion for each image is calculated according to $y_T=\frac{t}{(j-1)}$, where $t$ is a given time step and $j$ is the total number of time steps in the demonstration at hand. This neural architecture is novel as the timing features have never been used in this manner before. We are not interested in predicting the task completion proportion, but this branch will help to learn more features about the task during the model training.
\vskip5pt
\item {\it Attention-Based Encoder-Decoder} ({\bf Attention-RNN}): This architecture is an attention-based encoder-decoder using LSTM-based recurrent neural nets with Bahdanau attention \cite{bahdanau2016end}. While the encoder neural net generates features out of a history of images (10 in our case), the decoder predicts the sequence of labels. The inputs to the encoder are features extracted using the FCN model from input images. The inputs to the decoder are one hot encodings of the classification labels (see Fig.~\ref{models_RNN}).
\begin{figure}[!t]
    \centerline{\includegraphics[scale=0.56]{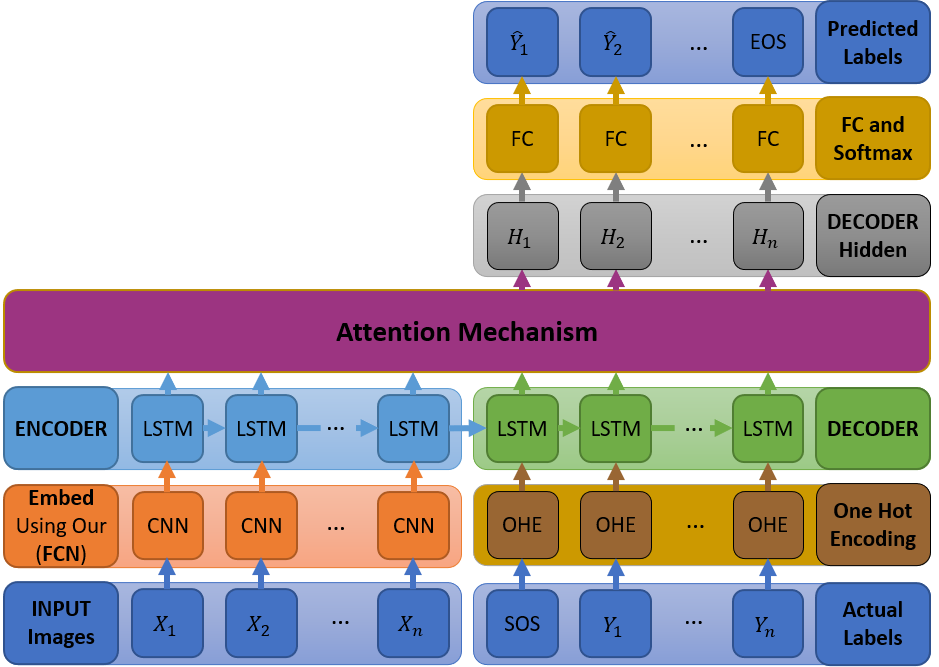}}
    \caption{Encoder-Decoder neural architecture}
    \label{models_RNN}
\end{figure}
\vskip5pt
\item {\it Transformer Network} ({\bf Transformer}): This architecture \cite{vaswani2017attention} has achieved state-of-the-art results in a number of applications, especially in natural language processing \cite{devlin2018bert,radford2019language,BrownMRSKDNSSAA20}. Similarly to {\bf Attention-RNN}, it uses the {\bf FCN} model for embedding input images and it also uses a history of 10 images. The decoder predicts the sequence of labels based on the multi-head attention layers and the features produced by our {\bf FCN} model (see Fig.~\ref{models_fig3}).
\vskip5pt
\item {\it Domain-Adversarial Neural Network} ({\bf DANN}): This architecture extends our \textbf{FCN} network by adding domain adaptation using the so-called Domain-Adversarial Training of Neural Networks (DANN) \cite{ganin2016domain,ganin2015unsupervised}. The main component of its domain discriminator path is a Gradient Reversal Layer (GRL), which reverses the gradient sign during backpropagation (see Fig.~\ref{models_fig2}). 
\vskip5pt
\item {\it Adversarial-Discriminative Domain Adaptation} ({\bf ADDA}): This is similar to \textbf{DANN}, the only difference is in the domain adaptation method. Here we used  Adversarial Discriminative Domain Adaptation (ADDA) \cite{tzeng2017adversarial}, which uses adversarial weights instead of GRL.
\vskip5pt
\item {\it Timing-Based and Domain-Based Fully Convolutional Net} ({\bf T-FCN-ADDA}): This architecture is similar to our \textbf{T-FCN} architecture, but it uses three paths instead of two: classification, timing, and domain. It is a novel neural architecture that combines the {\bf T-FCN} and {\bf ADDA} architectures above (see Fig.~\ref{models_fig2}). Similar to the T-FCN model, we are only interested in predicting the success probability using the classification path, but the timing and domain paths will help to learn more features about the task in hand.
\end{itemize}

\begin{figure}[!t]
    \centerline{\includegraphics[scale=0.466]{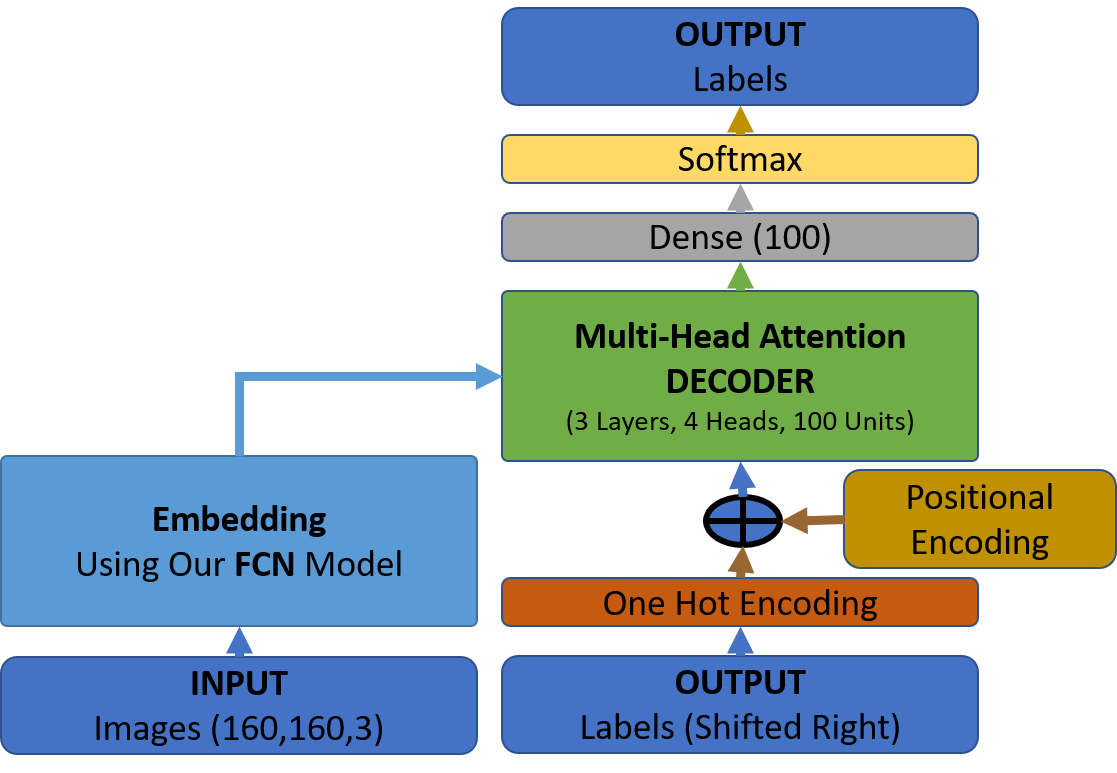}} 
    \caption{Transformer neural architecture}
    \label{models_fig3}
\end{figure}

\section{Evaluation}
\label{sec:evaluation}

\subsection{Experimental Setting}
\label{subsec:experiments}

	For each manipulation task, we split the demonstration data (images and classification labels) into a training set (80\%) and a validation set (20\%), while the unseen conditions data is used as a test set. In terms of loss functions, we used cross-entropy as a loss function for the success classifiers and the domain discriminator,  and the mean square error for the timing predictor\footnote{Summary of hyperparameters: batch size=16, epochs=100, optimizers=adam, and learning rate=0.001.}. This is due to the fact that the success predictors and domain discriminators are classification tasks, and the timing predictor is a regression task. All of the following experiments and tests were carried out for all fifteen tasks listed in Tables~\ref{tasks_datasets}, \ref{tasks_tab1}, and \ref{tasks_tab2}. Our experiments\footnote{PC specs: \textbf{CPU}: Intel i7-4770 @ 3.40GHz. \textbf{RAM}: 16GB. \textbf{GPU}: NVIDIA GeForce GTX 750 Ti 2GB.} focus on assessing the performance of success classification according to the following metrics: Classification Accuracy, Precision, Recall, F1-score, Area Under the Curve (AUC), Training Time, and Test Time. The latter two metrics refer to average times (in seconds) per image.
	
    \begin{table*}[!ht]
    \caption{Average performance results of our baseline and proposed neural architectures for task success classification applied to the Kitchen and MIME datasets (notation: \textbf{ACC}=Average Classification Accuracy, \textbf{AUC}=Area Under the Curve)}
    \begin{center}
    \begin{tabular}{|c||c|c||c|c|c|c|c||c|c|c|c|c|}
    \hline
                                         &                            &                    & \multicolumn{5}{c||}{Kitchen Dataset} & \multicolumn{5}{c|}{MIME Dataset}\\
    \textbf{Architecture}      & \textbf{Training} & \textbf{Test} & \textbf{ACC} & \textbf{Precision} & \textbf{Recall} & \textbf{F1} & \textbf{AUC} & \textbf{ACC} & \textbf{Precision} & \textbf{Recall} & \textbf{F1} & \textbf{AUC} \\
                                     & \textbf{Time} & \textbf{Time} & \textbf{} & \textbf{} & \textbf{} & \textbf{Score} & \textbf{} & \textbf{} & \textbf{} & \textbf{} & \textbf{Score} & \textbf{} \\
    \hline \hline
    \textbf{NASNet}    & 0.8044 & 0.8013 & 0.6190 & 0.1670 & 0.1612 & 0.1634 & 0.4192 & 0.8809 & 0.6186 & 0.6627 & 0.6165 & 0.7173 \\
    \textbf{FCN}   & 0.1373 & 0.0283 & 0.8240 & 0.9128 & 0.5244 & 0.6268 & 0.7586 & 0.9032 & 0.6211 & 0.8277 & 0.6917 & 0.7981 \\
    \textbf{T-FCN}  & 0.1921 & 0.0564 & 0.9131 & 0.9194 & 0.7716 & 0.8058 & 0.8636 & 0.8683 & 0.7694 & \textbf{0.8612} & 0.7660 & 0.8346 \\
    \textbf{Attention-RNN}   & 0.2133 & 0.0724 & 0.8380 & 0.8642 & 0.6570 & 0.6948 & 0.8776 & 0.8555 & 0.6870 & 0.6800 & 0.6278 & 0.7701 \\
    \textbf{Transformer}    & 0.2102 & 0.0681 & 0.8570 & 0.9130 & 0.5694 & 0.6338 & 0.7796 & 0.8576 & 0.6845 & 0.7443 & 0.6453 & 0.8084 \\
    \textbf{DANN}   & 1.9470 & 0.3250 & 0.9176 & 0.9052 & 0.7954 & 0.8334 & 0.8914 & 0.9410 & 0.8893 & 0.7438 & 0.7841 & 0.8372 \\
    \textbf{ADDA}  & 0.2082 & 0.0455 & 0.9577 & 0.9202 & 0.8980 & 0.9152 & 0.9300 & 0.9409 & 0.8668 & 0.7490 & 0.7612 & 0.8307 \\
    \textbf{T-FCN-ADDA} & 0.2209 & 0.0564 & \textbf{0.9733} & \textbf{0.9950} & \textbf{0.9052} & \textbf{0.9452} & \textbf{0.9642} & \textbf{0.9552} & \textbf{0.9429} & 0.8042 & \textbf{0.8397} & \textbf{0.9070} \\
    \hline
    \end{tabular}
    \label{results_tab1}
    \end{center}
    \end{table*}
	
    \begin{figure*}[h]
    \centerline{\includegraphics[scale=0.528]{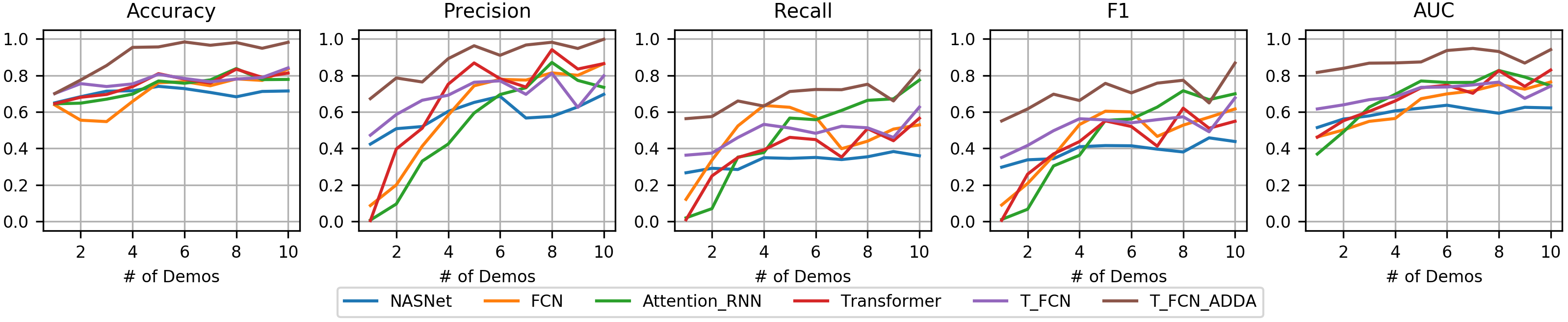}}
    \caption{Performance results for different amounts of demonstrations, from 1 to 10}
    \label{results_numdemos}
    \end{figure*}

\subsection{Experimental Results: Overview}
\label{subsec:results}

	The performance of our architectures---shown in Table~\ref{results_tab1}---is analysed in three groups. 
	First, we compare \textbf{NASNet} vs. \textbf{FCN} to observe their performance when trained using the data of five demonstrations and tested on unseen conditions with unseen distractor objects and different backgrounds. We also compare vanilla \textbf{FCN} vs. \textbf{T-FCN} to study the effects of using timing features.
        Second, we compare \textbf{Attention-RNN} vs. \textbf{Transformer} to study the sequential aspect of the manipulation tasks, and contrast their performance against the previous group. 
	Third, we investigate and compare two domain adaptation methods (\textbf{DANN} and \textbf{ADDA}) for mapping from the source domain (training) to the target domain (test=unseen conditions). In addition, we study the performance of \textbf{T-FCN-ADDA}---a combined model that uses both the timing and domain adaptation aspects on top of FCN learnt representations.
	
	
	\paragraph{Fully-Connected and Fully-Convolutional Classifiers}
	Table~\ref{results_tab1} shows that \textbf{T-FCN} outperforms \textbf{NASNet} and \textbf{FCN} across most of the metrics. The difference in performance according to F1-Score is more stark than the other metrics (ACC and AUC). It can also be noted that \textbf{NASNet} is computationally more expensive, not only during training but also at test time. Assuming the classification of one million images, while \textbf{FCN} would require 7.9 hours, \textbf{T-FCN} would require 15.7 hours, and \textbf{NASNet} would require 225.6 hours. Thus, \textbf{T-FCN} is not only the most accurate in this group, but it is also substantially faster at training and prediction times than \textbf{NASNet}---even when it is a slower predictor than \textbf{FCN}.
    
	\paragraph{Seq-to-Seq Classifiers}
	Table~\ref{results_tab1} shows that \textbf{Transformer} outperforms \textbf{Attention-RNN} according to ACC, but not according to F1-Score and AUC. Given that our dataset is imbalanced, we could rely on F1-Scores instead of ACC; but we found that we can rely on ACC. We found strong correlations between ACC and F1, ACC and AUC, and F1 and AUC, obtaining Pearson correlation coefficients of 0.79, 0.64, and 0.91, respectively. The fact that both classifiers in this group are outperformed by the best classifier of the first group (\textbf{T-FCN}) (across all metrics), suggests that further research on this type of architectures is needed to integrate sequential aspects into the top success classifier.

	\paragraph{Domain Adaptation Classifiers}
	Table~\ref{results_tab1} shows that the use of domain adaptation techniques (\textbf{DANN} and \textbf{ADDA}) helps to achieve better results than the architectures in the previous groups. Although \textbf{ADDA} outperforms \textbf{DANN} across most metrics, the combination of \textbf{T-FCN} and \textbf{ADDA} (i.e. \textbf{T-FCN-ADDA}) achieves the best classification results. While  \textbf{T-FCN} and \textbf{ADDA} achieve average F1-scores of (80.6\%, 76.6\%) and (91.5\%, 76.1\%) across tasks for both datasets (respectively), \textbf{T-FCN-ADDA} achieves an average F1-scores of (94.5\%, 84\%) in both datasets. Regarding prediction times, \textbf{T-FCN-ADDA} is comparable to \textbf{T-FCN}---the third fastest in our neural architectures, just after vanilla {\bf FCN} and  {\bf ADDA}. This is due to the fact that the domain discriminator is not used at test time. 
	
\subsection{Experimental Results: Analysis}

A natural question to ask is ``How many demonstrations are needed to train task success classifiers?'. We analysed the answer to this question and found that there is a small improvement in performance when increasing the number of demos used for training from 5 to 10 demos (see Fig.~\ref{results_numdemos}). This suggests that the slightly small improvement in performance is not worth the double efforts from human demonstrators. This result justifies our selection regarding the use of 5 demonstrations to train our models.

    
From Table~\ref{results_confmat} it can be noted that the true positive rates (TPR, or recall) and false positive rates (TNR) are higher for \textbf{T-FCN-ADDA} than its counterparts. This is clearly the case on average for all tasks in both datasets, but also the case on most individual tasks. These results show evidence that our top classifier (proposed) is more reliable than the baselines.  

\begin{figure}[!tbp]
    \centerline{\includegraphics[scale=0.38]{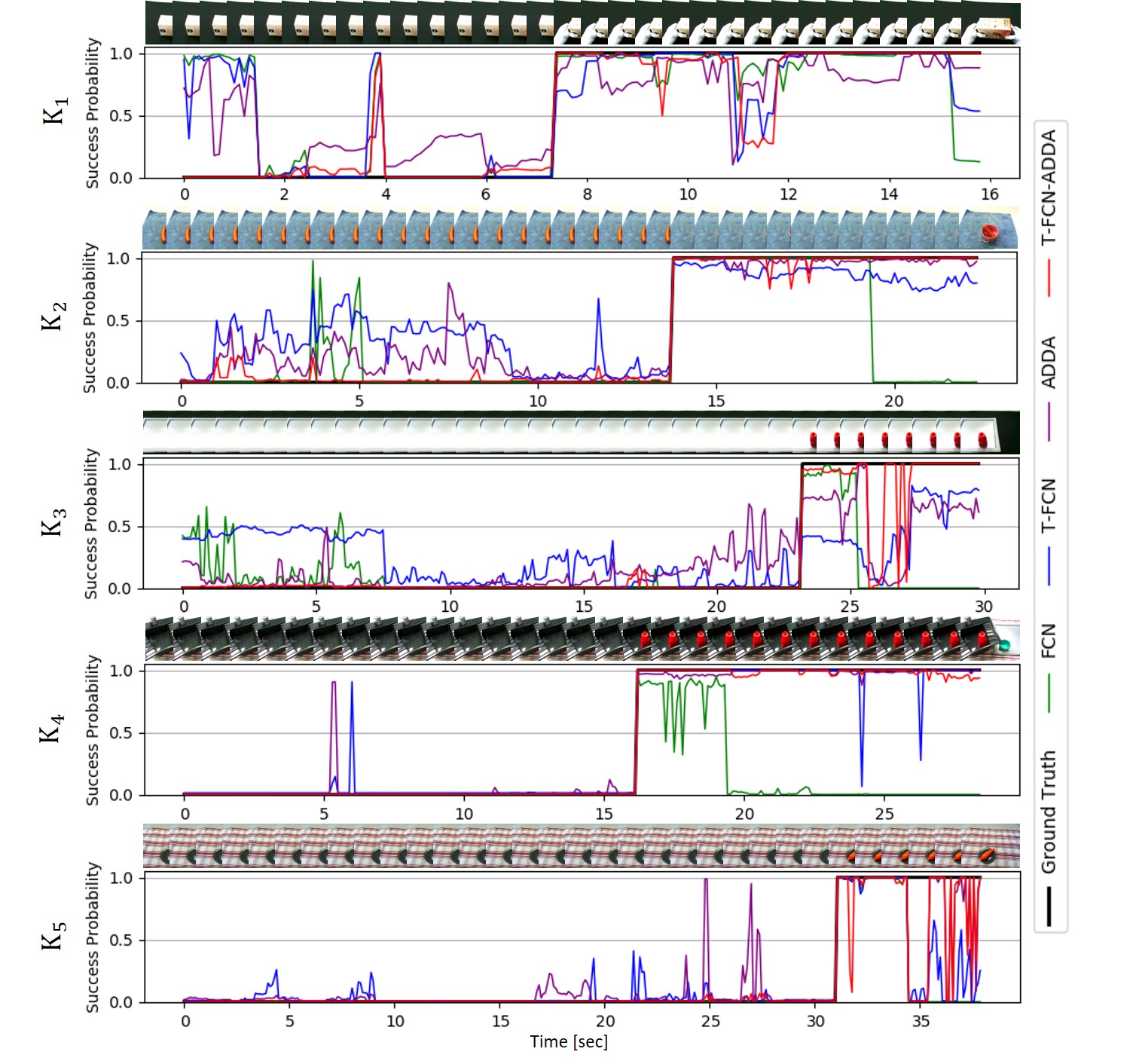}}
    \caption{Illustration of predicted probabilities generated by {\bf T-FCN-ADDA} and three baselines on the Kitchen dataset.}
    \label{results_fig2}
    \end{figure}

\begin{figure*}[!tbp]
  \centering
  {\includegraphics[width=0.517\textwidth]{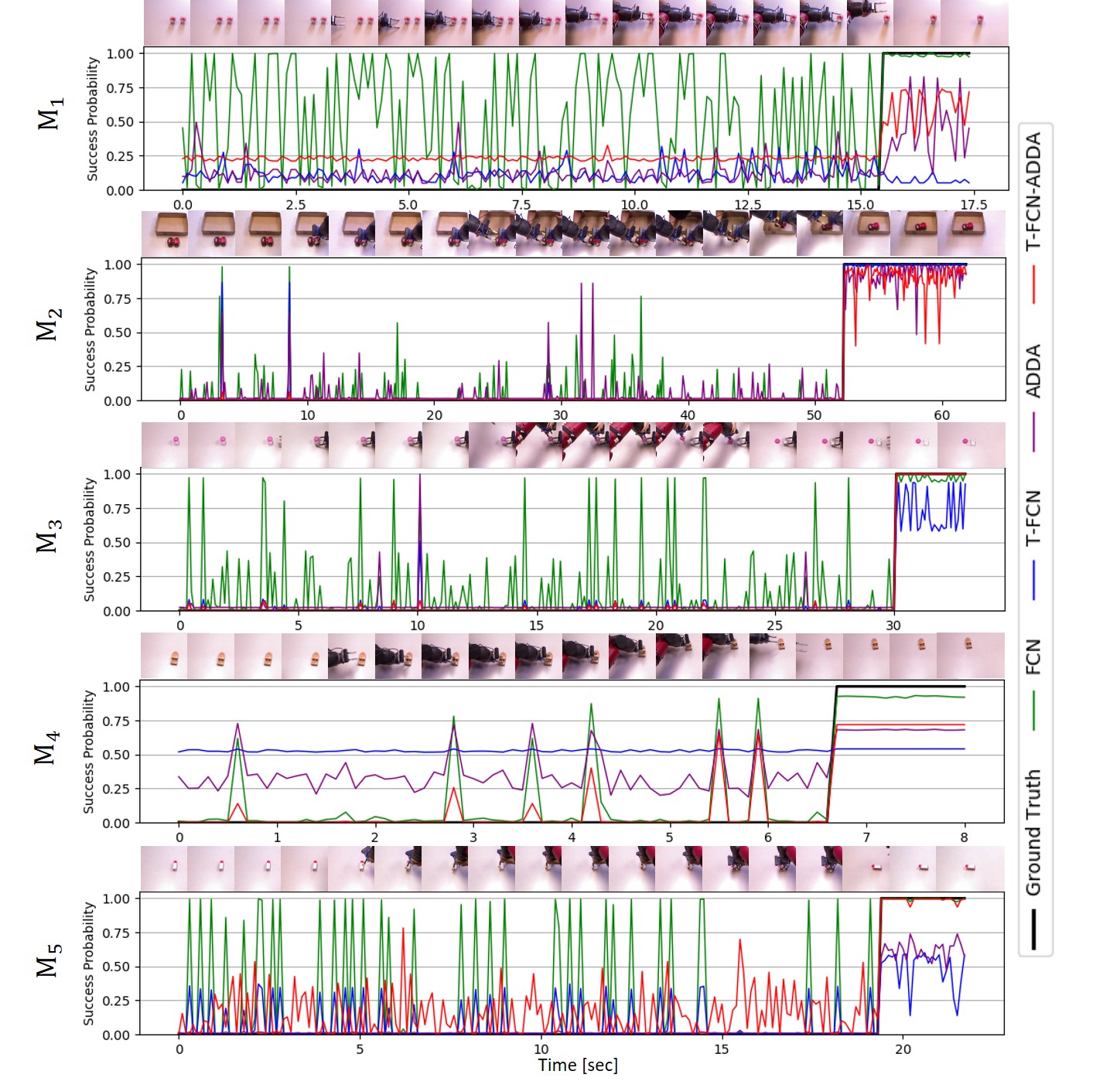}\label{results_probs_MIME1}}
  \hfill
  \hskip-30pt
  {\includegraphics[width=0.517\textwidth]{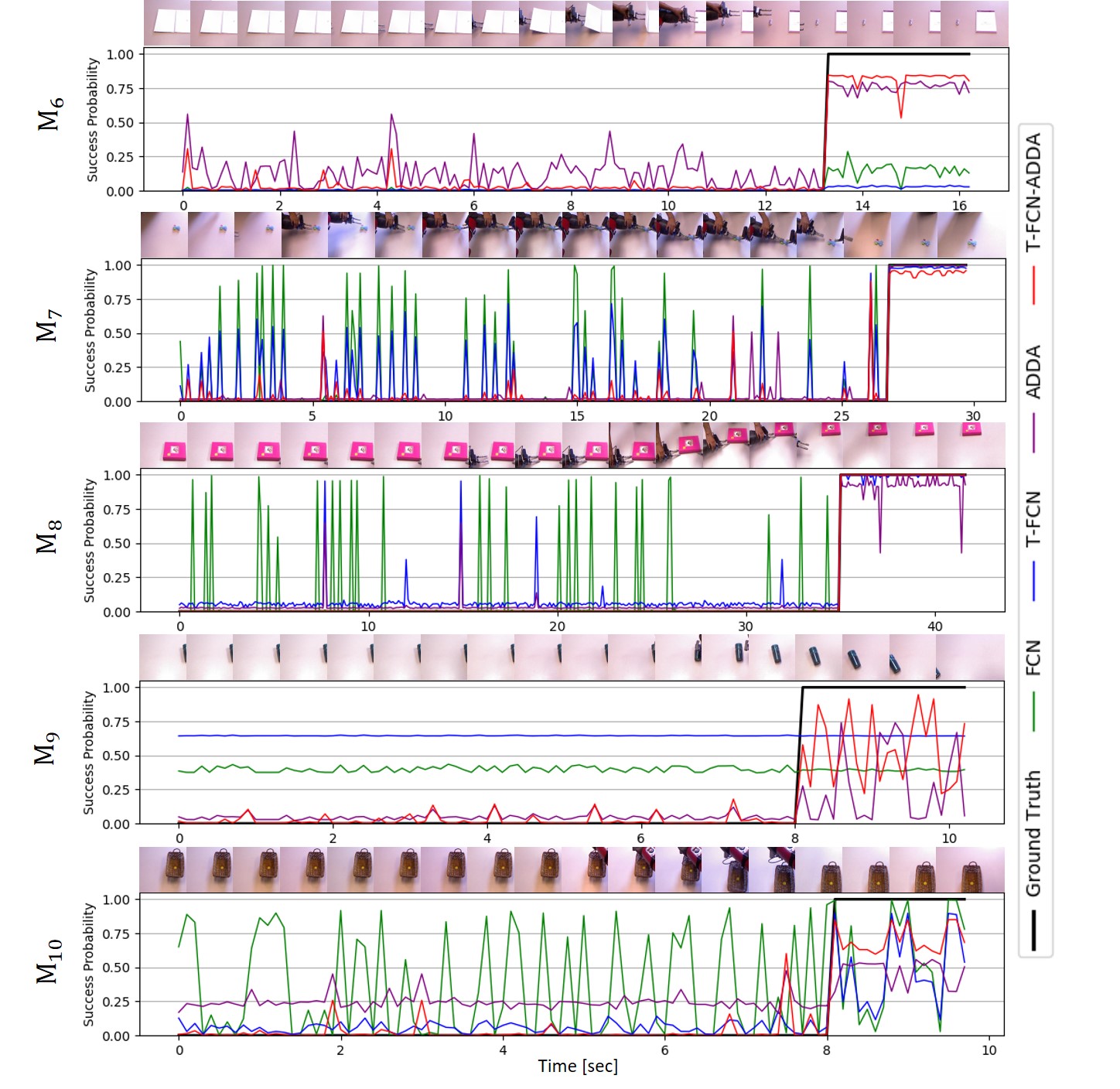}\label{results_probs_MIME2}}
  \caption{Illustration of predicted probabilities generated by {\bf T-FCN-ADDA} and three baselines on the MIME dataset.}
  \label{results_probs_MIME}
\end{figure*}

To illustrate the performance of our task success classifiers, Figs.~\ref{results_fig2} and~\ref{results_probs_MIME} show the predicted success probabilities for example demonstrations of our top classifier (\textbf{T-FCN-ADDA}) and three baseline classifiers (\textbf{FCN}, \textbf{T-FCN}, and \textbf{ADDA}). A visual inspection shows that our top classifier is closer to the ground truth than the baselines in both datasets.

    \begin{table}[!t]
    \scriptsize
        \begin{center}
    \begin{tabular}{c|c|c|c|c|c|c|c|c}
    \hline
    \multirow{2}{*}{Task} & \multicolumn{2}{c|}{\bf FCN} & \multicolumn{2}{c|}{\bf T-FCN} & \multicolumn{2}{c|}{\bf ADDA} & \multicolumn{2}{c}{\bf T-FCN-ADDA} \\
    {\bf } & TPR & TNR & TPR & TNR & TPR & TNR & TPR & TNR \\
    \hline \hline
    \tiny K$_1$ & 0.903	& 0.823	& 0.892	& 0.906	& 0.971	& 0.933	& 0.911	& 0.975 \\
    \tiny K$_2$ & 0.817	& 0.933	& 1.000	& 0.835	& 1.000	& 0.945	& 1.000	& 1.000 \\
    \tiny K$_3$ & 0.832	& 0.808	& 0.840	& 0.800	& 0.927	& 0.907	& 0.967	& 1.000 \\
    \tiny K$_4$ & 0.633	& 1.000	& 0.988	& 0.992	& 1.000	& 0.984	& 1.000	& 1.000 \\
    \tiny K$_5$ & 0.899	& 1.000	& 0.907	& 1.000	& 0.951	& 0.946	& 0.942	& 1.000 \\
    \hline
    \tiny M$_1$ & 1.000	& 0.345	& 0.934	& 0.905	& 0.908	& 0.684	& 0.956	& 0.967 \\
    \tiny M$_2$ & 0.999	& 0.940	& 0.991	& 0.985	& 0.997	& 0.969	& 0.984	& 1.000 \\
    \tiny M$_3$ & 0.999	& 0.609	& 0.996	& 0.977	& 1.000	& 0.985	& 0.999	& 0.985 \\
    \tiny M$_4$ & 1.000	& 0.825	& 1.000	& 0.571	& 1.000	& 0.800	& 1.000	& 0.870 \\
    \tiny M$_5$ & 0.998	& 0.508	& 0.938	& 0.599	& 0.999	& 0.956	& 0.966	& 0.912 \\
    \tiny M$_6$ & 0.938	& 1.000	& 0.898	& 1.000	& 0.967	& 0.829	& 0.987	& 0.964 \\
    \tiny M$_7$ & 0.995	& 0.593	& 1.000	& 0.645	& 0.995	& 0.911	& 0.996	& 0.960 \\
    \tiny M$_8$ & 0.978	& 0.809	& 0.995	& 0.861	& 0.996	& 0.985	& 0.998	& 0.988 \\
    \tiny M$_9$ & 0.744	& 0.000	& 0.000	& 0.256	& 0.779	& 0.889	& 0.781	& 1.000 \\
    \tiny M$_{10}$ & 0.979	& 0.357	& 0.968	& 0.490	& 0.896	& 0.667	& 0.940	& 0.783 \\
    \hline \hline
    Avg. & 0.914	& 0.703	& 0.890	& 0.788	& 0.959	& 0.893	& {\bf 0.962}	& {\bf 0.960} \\
    Std. & 0.112	& 0.296	& 0.251	& 0.226	& 0.061	& 0.104	& {\bf 0.057}	& {\bf 0.062} \\
    \hline
    \end{tabular}
    \caption{Performance of our {\bf T-FCN-ADDA} classifier and three baselines. Notation: TPR=TP Rate, TNR=FP Rate, TPR=TP/(TP+FN), TNR=TN/(TN+FP), TP=True Positives, TN=True Negatives, FP=False Positives, FN=False Negatives}
    \label{results_confmat}
    \end{center}
    \end{table}
    
\section{Concluding Remarks}
This paper studies the problem of reward learning for robotic manipulation via neural task success classifiers, where the aim is to find out whether it is possible to train accurate task success classifiers from few demonstrations. Our study is a step in the direction of autonomous robot skill acquisition, where a human demonstrator shows a humanoid robot how to carry out a novel task. Our experiments are focused on predicting (probabilistically) whether the robot has achieved its task or not. We carry out a comprehensive comparison of three types of neural architectures: (i) fully-connected and fully-convolutional neural nets, (ii) sequence-to-sequence learning, and (iii) domain adaptation learning.

Our experiments use two datasets containing images on different tasks and with varied backgrounds and distractor objects. Experimental results reveal that 5 human demonstrations is a good compromise between effort and performance. They also show that {\bf T-FCN} is the best architecture from the first group; {\bf Transformer} is the best from the second group according to classification accuracy; {\bf Attention-RNN} is the best according to F1-score; and {\bf T-FCN-ADDA} is the best architecture not only from the third group but from the three groups of classifiers. {\bf T-FCN-ADDA} is a novel solution that combines the best architecture from group one and the best domain adaptation classifier. It achieves high performance across all tasks in the Kitchen and MIME datasets with average classification accuracies of {\bf 97.3\%} and {\bf 95.5\%} in those datasets, while vanilla {\bf FCN} models only reach {\bf 82.4\%} and {\bf 90.3\%}, respectively. 

Future works include investigating reward learning in more complex tasks than those attempted here, training robots to carry out manipulation tasks using the proposed classifiers, and studying their application to other robot platforms. 


%


\bibliographystyle{IEEEbib}
\bibliography{am-ijcnn2021}

\end{document}